\pdfoutput=1 
\documentclass{nature}
\usepackage{lineno}
\usepackage{underscore}
\usepackage{multirow}
\usepackage{indentfirst}

\usepackage{ccaption}
\usepackage{subcaption}
\usepackage{graphicx}
\usepackage{lipsum}
\usepackage{ragged2e}
\usepackage{hyperref}% to split url multiple lines

\usepackage{cmbright}
\usepackage{filecontents}
\usepackage{amsmath}
\renewcommand\refname{References}

\newcounter{lastnote}

\usepackage{comment}
\usepackage{cite}
\usepackage{adjustbox} % to adjust the size of table

\makeatletter
\newenvironment{figurehere}
{\def\@captype{figure}}
{}
\makeatletter

\usepackage{xcolor}
\definecolor{sared}{rgb}{0.69, 0,0}
\title{\begin{center}  A Unified Memory Perspective for Probabilistic Trustworthy AI \end{center} }

\author
{ 
\centering
\large{
Xueji Zhao$^{1*}$, 
Likai Pei$^{1*}$, 
Jianbo Liu$^{1}$, 
Kai Ni$^{1}$, 
Ningyuan Cao$^{1\dagger}$\\
}
\vspace{5ex}
\normalsize{
$^{1}$University of Notre Dame, Notre Dame, IN 46556, USA\\
$^*$These authors contributed equally to this work\\
$^\dagger$To whom correspondence should be addressed \\
Email: ncao@nd.edu\\
}
}

\begin{document}
\flushbottom
\maketitle

\begin{abstract}
\vspace{3ex}
Trustworthy artificial intelligence increasingly relies on probabilistic computation to achieve robustness, interpretability, security and privacy. In practical systems, such workloads interleave deterministic data access with repeated stochastic sampling across models, data paths and system functions, shifting performance bottlenecks from arithmetic units to memory systems that must deliver both data and randomness. Here we present a unified data-access perspective in which deterministic access is treated as a limiting case of stochastic sampling, enabling both modes to be analyzed within a common framework. This view reveals that increasing stochastic demand reduces effective data-access efficiency and can drive systems into entropy-limited operation. Based on this insight, we define memory-level evaluation criteria, including unified operation, distribution programmability, efficiency, robustness to hardware non-idealities and parallel compatibility. Using these criteria, we analyze limitations of conventional architectures and examine emerging probabilistic compute-in-memory approaches that integrate sampling with memory access, outlining pathways toward scalable hardware for trustworthy AI.

\end{abstract}


%\noindent Please note: Abbreviations should be introduced at the first mention in the main text – no abbreviations lists. Suggested structure of main text (not enforced) is provided below.

\section*{\textcolor{sared}{\large Introduction}}

Artificial intelligence systems are increasingly deployed in high-stakes settings, including medical decision-making~\cite{metwallyPredictionMetabolicSubphenotypes2025}, autonomous platforms~\cite{automated_driving_nn} and robotic agents~\cite{science_robotics_uhang}. In these applications, reliable operation requires more than accurate prediction. Systems must also quantify uncertainty~\cite{isscc_shot_noise}, explain their decisions~\cite{nutiExplainableBayesianDecision2021} and protect sensitive information~\cite{davisInSituPrivacyMixedSignal2024}. Uncertainty is therefore not a peripheral disturbance but a structural aspect of the problem, arising from noisy inputs, evolving environments and imperfect or adversarial models. As a result, modern AI systems increasingly rely on probability and randomness to support trustworthy operation.
This shift fundamentally changes the nature of computation. Beyond processing stored deterministic data, systems must now continuously generate, transport and consume stochastic information. As probabilistic computation becomes pervasive, randomness itself emerges as a first-class computational resource, placing new and growing demands on the underlying hardware substrate, particularly on memory systems that must support both data access and stochastic sampling~\cite{writing_reading_application}.

Prior work has examined probabilistic computation from perspectives including probabilistic algorithms~\cite{ghahramani_probabilistic_2015}, device-level randomness generation~\cite{distribution_conversion}, and secure compute-in-memory systems~\cite{wang_safe_2024}. In this Perspective, we instead focus on the interaction between probabilistic computation and memory access. From this viewpoint, stochastic sampling can be interpreted as a form of generalized data access: probabilistic memory access returns a sample drawn from a distribution associated with a memory location, while deterministic access corresponds to the limiting case of zero variance. This unified abstraction places random number generation and deterministic memory operations within a common framework.

Building on this perspective, we make three main contributions. First, we introduce a unified probabilistic memory abstraction that enables deterministic and stochastic operations to be analyzed within a single framework. Second, we identify a fundamental scaling mismatch among compute throughput, memory bandwidth and entropy generation, showing that increasing stochastic demand can shift systems into entropy-bound regimes, or "entropy wall". Third, we examine architectural trade-offs across conventional von Neumann systems and emerging probabilistic compute-in-memory (CIM) approaches, and outline cross-layer opportunities spanning devices, circuits, architectures and software abstractions for probabilistic computation.

\section*{\textcolor{sared}{\large   A Memory-Centric View of Probabilistic Workloads}}

\noindent\textbf{Co-evolution of probabilistic computation and hardware}

Stochastic sampling has co-evolved with computing platforms for decades, enabling increasingly complex probabilistic algorithms as illustrated in Figure.~\ref{fig:overview}. In modern systems, advances in hardware have supported methods ranging from classical Monte Carlo estimation~\cite{NE-endurance} to particle filtering~\cite{particle_filtering} and large-scale decision-making frameworks such as Monte Carlo Tree Search (MCTS)~\cite{mc_tree_search}. At the same time, random number generation has progressed from software-based pseudo-random methods~\cite{bernstein2008chacha} to high-throughput hardware entropy sources~\cite{isscc_shot_noise}. Crucially, this co-evolution has been accompanied by a steady increase in stochastic sampling demand per task~\cite{silver_mastering_2017}, placing growing pressure on the underlying computing substrate.

\begin{figurehere}
   \centering
    \includegraphics[width=1\linewidth]{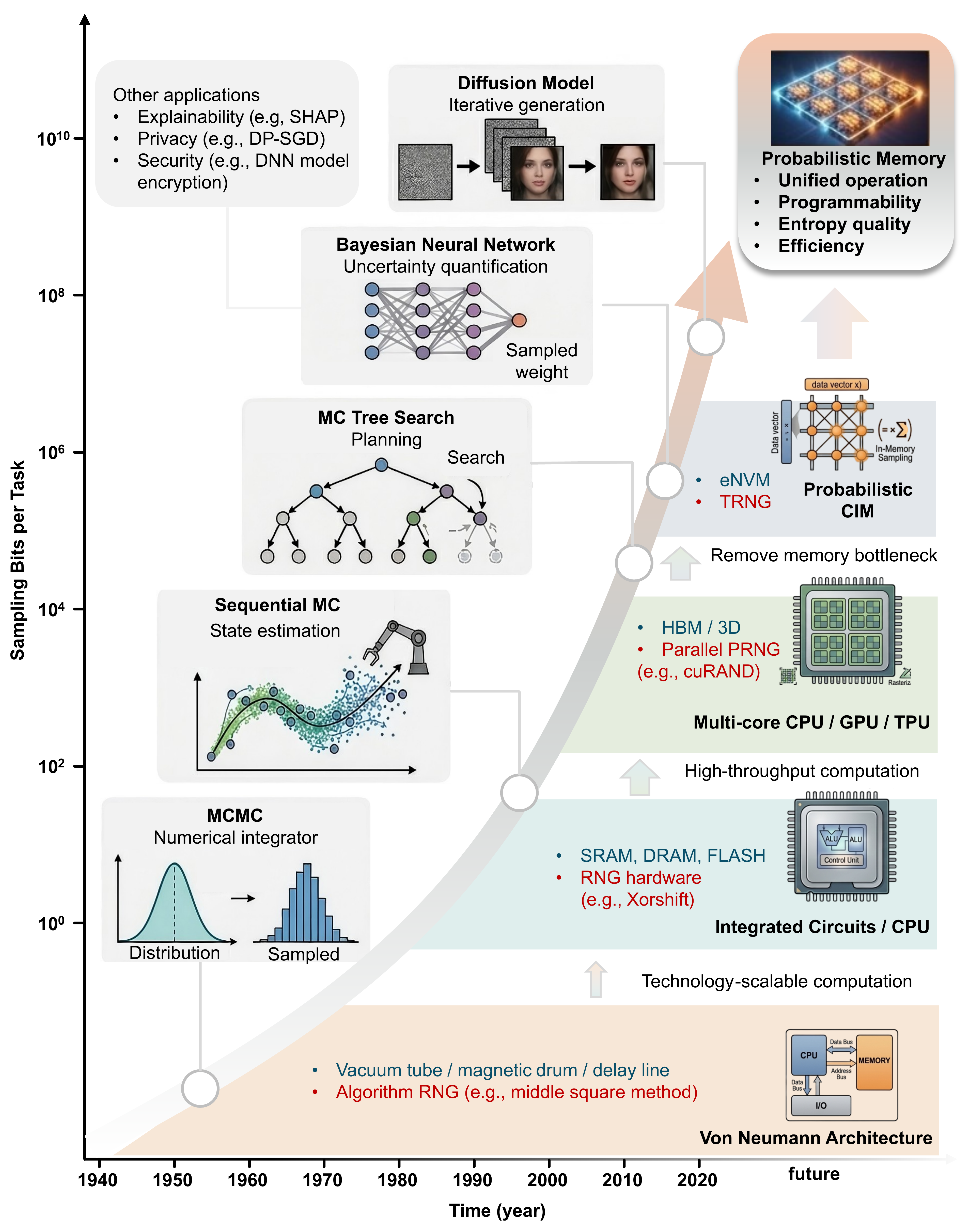}
    \captionsetup{parbox=none} % for caption to be split
    \caption{\textit{\textbf{Co-evolution of probabilistic computing workloads and hardware architectures.} Over eight decades, entropy requirements per task have scaled exponentially, driven by the transition from classical sampling to modern uncertainty quantification and generative AI. The left panel illustrates the algorithmic advancement from early Markov Chain Monte Carlo (MCMC) to complex Bayesian Neural Networks and Diffusion Models. Correspondingly, the right panel depicts the shift in computing paradigms to overcome the "memory wall," moving from Von Neumann architectures to high-throughput GPUs and emerging probabilistic computing-in-memory (p-CIM). This trajectory converges toward \emph{probabilistic memory} (top right), which leverages intrinsic device randomness to provide unified operations and high-quality entropy with superior energy efficiency.}}
    \label{fig:overview}
\end{figurehere}

This trend not only persists but intensifies in contemporary trustworthy AI workloads, where stochastic operations are deeply integrated into model execution. Bayesian and uncertainty-aware models rely on repeated sampling to characterize predictive uncertainty~\cite{peiUncertaintyawareRoboticPerception2025a}, generative models such as diffusion models produce outputs through iterative stochastic processes~\cite{cao_survey_2023}, interpretability methods leverage randomized perturbations~\cite{ribeiro_lime_2016}, and privacy-preserving systems introduce controlled noise injection~\cite{davisInSituPrivacyMixedSignal2024}. As a result, stochastic sampling is no longer an auxiliary procedure but a dominant component of computation. In many cases, the number of stochastic samples required per inference or decision can approach or exceed the number of deterministic data accesses~\cite{jun2014memory}, fundamentally altering the balance between computation and data movement.

This shift exposes new system-level bottlenecks. While compute throughput has increased dramatically~\cite{google_tpu} and memory bandwidth has improved more modestly~\cite{memory_wall_wulf}, the throughput of random number generation and entropy sources has scaled much more slowly~\cite{trng_throughput}. As stochastic demand increases, effective system throughput becomes bounded by entropy generation rather than computation or memory bandwidth. In this regime, performance scales with the rate of entropy generation and delivery. This emerging ``entropy wall'' extends the classical memory wall and introduces a new dimension to system design.

To address data movement challenges, CIM architectures have been explored to bring computation closer to memory~\cite{cim_nature_sun_2023}. More recently, probabilistic compute-in-memory (p-CIM) approaches integrate stochastic functionality directly within memory structures by leveraging intrinsic device randomness~\cite{peiUncertaintyawareRoboticPerception2025a}, enabling in-situ sampling and probabilistic operations. These architectures aim to reduce both sampling overhead and data movement for probabilistic workloads such as Bayesian neural networks and uncertainty-aware models~\cite{ICCAD}. Figure~\ref{fig:overview} summarizes this co-evolution across probabilistic algorithms, computing platforms, random number generation and memory systems.

\noindent\textbf{A unified memory perspective for probabilistic computation} 

\noindent Stochastic sampling rarely appears in isolation. Instead, it is repeatedly interleaved with parameter reads (such as Gaussian mean and variance~\cite{rezende2014stochastic}), feature access (as in variational autoencoders~\cite{vae}), and model evaluation (as in Bayesian neural networks~\cite{ICCAD, isscc_shot_noise}). As probabilistic algorithms become increasingly integrated with neural networks, sampling and memory access become tightly coupled.
This observation motivates a unified perspective, independent of specific hardware implementations, as illustrated in Figure.~\ref{fig:bottleneck}(a,b). Conventional memory systems are designed for deterministic reads that return fixed stored values, whereas probabilistic workloads require sampling from distributions associated with parameters, latent variables, or stochastic processes. From this viewpoint, deterministic memory access can be interpreted as a limiting case of sampling with vanishing variance. Deterministic reads, stochastic sampling, and random number generation can thus be unified within a common data-access abstraction.

\begin{figurehere}
\vspace{3ex}
   \centering
    \includegraphics[width=0.95\linewidth]{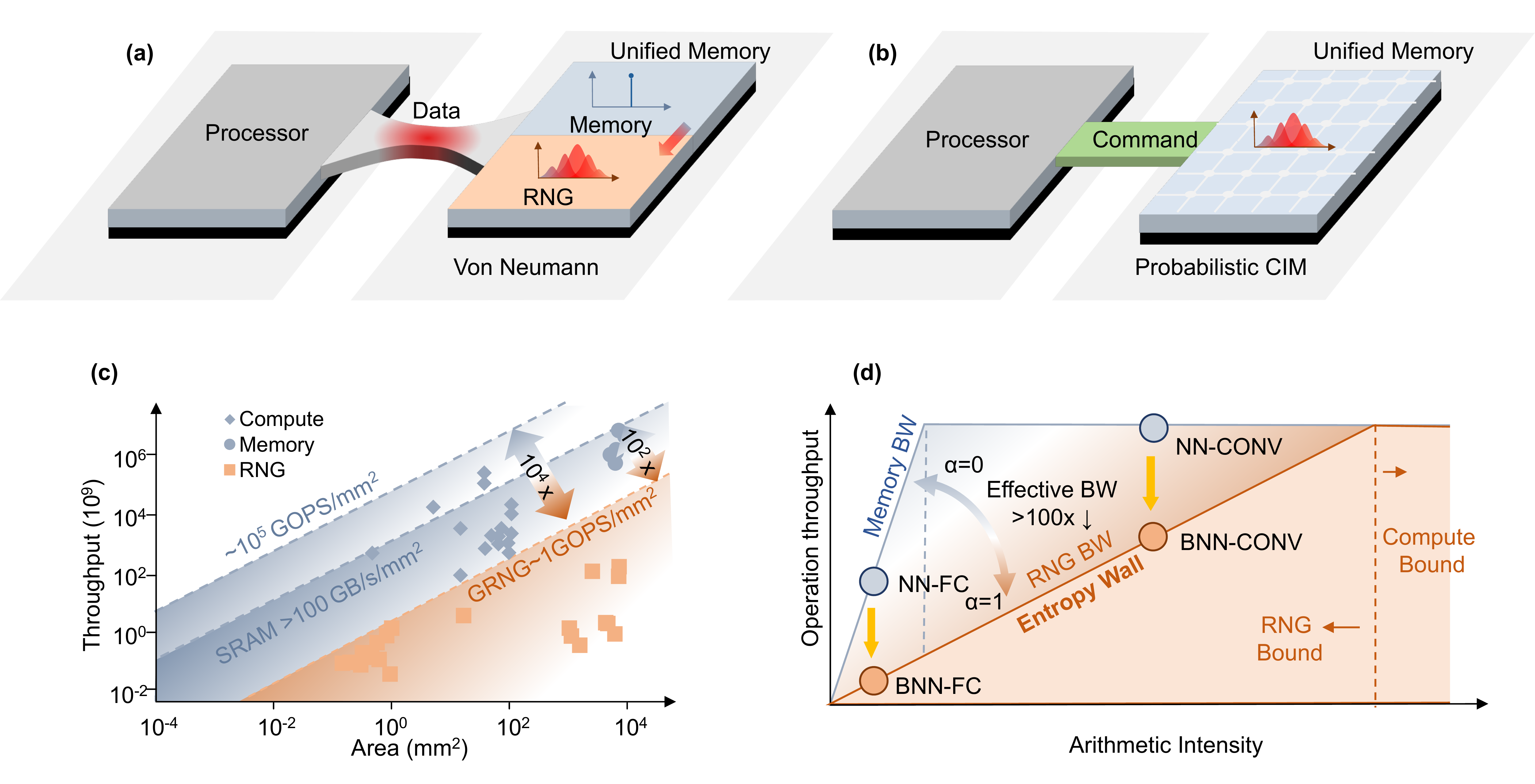}
    \captionsetup{parbox=none} % for caption to be split
  \caption{
\textbf{From memory wall to entropy wall: a unified view of probabilistic data access.}
\textit{\textbf{a,} Conventional von Neumann architectures separate deterministic memory access and random number generation (RNG), forcing stochastic data to share interconnect pathways and creating data and entropy bottlenecks. 
\textbf{b,} Probabilistic compute-in-memory (p-CIM) integrates sampling within memory, enabling unified access to deterministic data and probabilistic distributions. 
\textbf{c,} Scaling trends reveal a pronounced mismatch: while compute and memory scale aggressively (e.g., $\sim$10 TOPS/mm$^2$, $>100$ GB/s/mm$^2$), entropy generation (e.g., GRNG) lags by orders of magnitude. 
\textbf{d,} Unified throughput model versus arithmetic intensity: as the probabilistic data ratio $\alpha$ increases, workloads shift from memory-bound ($\alpha \rightarrow 0$) to entropy-bound ($\alpha \rightarrow 1$). Even moderate stochastic demand can reduce effective bandwidth ($>100\times$), pushing workloads such as Bayesian neural networks into entropy-limited regimes.}}
    \label{fig:bottleneck}
\end{figurehere}

Building on this abstraction, we derive a system-level performance model. Rather than treating deterministic memory access and stochastic sampling as separate operations, they are viewed as components of a unified data-access process. System throughput is therefore determined by compute throughput ($\pi$) and a unified data-access throughput ($\beta$) that jointly captures deterministic data access and entropy generation.

We define a probabilistic data ratio $\alpha \in [0,1]$ as the fraction of stochastic (entropy-driven) accesses relative to total data access. Under this definition, the effective data-access throughput can be expressed as:
\begin{equation}
\frac{1}{\beta} = \frac{\alpha}{\beta_{\mathrm{rand}}} + \frac{1-\alpha}{\beta_{\mathrm{data}}},
\label{eq:alpha}
\end{equation}
where $\beta_{\mathrm{rand}}$ and $\beta_{\mathrm{data}}$ denote the entropy generation throughput and deterministic data access throughput, respectively. The overall system throughput $\Phi$ is then approximated as:
\begin{equation}
\Phi \approx \min\left(\pi,\; AI \cdot \beta\right),
\label{eq:roofline}
\end{equation}
where $AI$ denotes the arithmetic intensity, defined as the number of operations per total data movement, including both deterministic accesses and stochastic sampling.

This formulation captures a continuous transition between operating regimes. In the limiting case $\alpha \rightarrow 0$, performance reduces to a conventional data-bound regime governed by deterministic memory access. In contrast, as $\alpha \rightarrow 1$, the system approaches an entropy-bound regime dominated by stochastic generation throughput. Increasing $\alpha$ therefore shifts the bottleneck from memory bandwidth to entropy supply, unifying deterministic and probabilistic workloads within a single roofline-like framework.

This transition is highly asymmetric in practical systems. In typical von Neumann digital architectures, entropy generation throughput  is orders of magnitude lower than deterministic data access throughput, as randomness is typically produced by narrow, specialized circuits or peripheral units rather than wide, high-throughput memory interfaces. This reveals a disproportionate sensitivity: the system is pushed into the entropy-bound regime even when stochastic demand is seemingly negligible. For example, given that the throughput gap between memory access and entropy generation typically exceeds two orders of magnitude, even a small stochastic fraction (e.g., $\alpha \approx 1\%$) can induce entropy-limited behavior. This throughput disparity and the resulting "entropy wall" are illustrated in Fig.~\ref{fig:bottleneck}(c,d) and discussed in details in next section.

%In many von Neumann systems, randomness shares interconnect and memory pathways with deterministic data, introducing bandwidth contention and compounding both computation and data-movement overheads, thereby giving rise to an “entropy wall” alongside the classical memory wall. This unified formulation treats deterministic access and stochastic sampling as a single data-access primitive, where entropy is integrated into the access pathway rather than supplied externally. As a result, system performance is governed not only by data movement, but also by the availability and delivery of entropy as a computational resource.

\noindent\textbf{Architectural implications for probabilistic memory}

\noindent
Building on the unified data-access model (Eqs.~\ref{eq:alpha}--\ref{eq:roofline}), where effective throughput is jointly limited by memory bandwidth and entropy generation with a bottleneck that shifts with the probabilistic data ratio $\alpha$, one can infer the properties that a unified memory primitive should provide to sustain performance across operating regimes.

First, since deterministic and stochastic accesses are unified through $\alpha$, memory systems should support \textit{unified deterministic and probabilistic primitives}, treating deterministic access as the zero-variance limit of sampling within a common pathway. 

Second, as different workloads correspond to different distributions and $\alpha$ values, systems should enable \textit{reconfigurable distribution shaping} to accommodate diverse and dynamically varying probabilistic workloads. 

Third, because entropy generation directly impacts effective throughput in the entropy-bound regime, systems should maintain \textit{statistical fidelity and robustness}, as bias, correlation, and temporal drift can propagate through sampling and degrade inference quality.

Finally, as increasing $\alpha$ shifts the system toward an entropy-limited regime, these capabilities should be achieved with \textit{high scaling efficiency}, requiring sampling throughput, energy efficiency, and memory density to scale alongside computational demand.

Collectively, these model-driven requirements move beyond architectures that treat randomness as an auxiliary function, instead positioning stochastic sampling as a first-class data-access primitive co-designed with deterministic access for scalable trustworthy AI.

\section*{\textcolor{sared}{\large  Von Neumann Structural Limitation}}

\noindent\textbf{Random number generation in von Neumann systems}

\noindent In conventional von Neumann architectures, stochastic computation relies on explicit random number generation pipelines that suffer from systemic data-movement bottlenecks. Random numbers are typically produced by pseudo-random generators, such as linear congruential generators~\cite{knuth1997art}, Mersenne Twister~\cite{matsumotoMersenneTwister623dimensionally1998}, and then transformed into target distributions through additional computation. This separation provides flexibility and high statistical quality, but incurs substantial overhead in the unified data-access model. Sampling from common distributions, such as Gaussian, requires additional arithmetic, lookup tables, or rejection-based methods (e.g., Box--Muller, Ziggurat, Wallace~\cite{thomasGaussianRandomNumber2007a}), increasing latency, control complexity, and data movement. As a result, stochastic sampling introduces significantly higher effective cost than deterministic data access, especially as sampling demand increases.

\noindent\textbf{When sampling becomes the bottleneck}

\noindent In von Neumann systems, deterministic data access and stochastic sampling are supported by fundamentally different hardware pathways, as illustrated in Figure.~\ref{fig:bottleneck}(a). While deterministic data is delivered through high-bandwidth memory systems, stochastic data must be generated through separate RNG pipelines, resulting in a structural mismatch in data-access throughput.
This mismatch is further amplified by divergent scaling trends Figure.~\ref{fig:bottleneck}(c). State-of-the-art RNG implementations achieve throughput densities on the order of 1 GSa s$^{-1}$ mm$^{-2}$~\cite{malikGaussianRandomNumber2016}, whereas on-chip memory bandwidth can exceed $10^{2}$ GB s$^{-1}$ mm$^{-2}$~\cite{luoBenchmarking2024}, and compute fabrics can reach $10^{4}$--$10^{5}$ GOPS mm$^{-2}$~\cite{khwa14216nm216kb2025}. This growing disparity places entropy generation on a fundamentally different scaling trajectory from memory and computation. 

As the probabilistic data ratio $\alpha$ increases, this imbalance does not merely shift the operating point along a fixed roofline, but reshapes the effective system constraint. Stochastic sampling reduces the effective data-access throughput, lowering achievable performance even at the same arithmetic intensity.
As illustrated in Fig.~\ref{fig:bottleneck}(d), increasing $\alpha$ effectively compresses the memory bandwidth ceiling. Workloads that are originally memory-bound become further constrained by reduced effective bandwidth, while workloads that are originally compute-bound—such as convolution with high data reuse—can transition into memory-bound regimes. For example, in Bayesian neural networks, stochasticity is introduced at the parameter level. Each weight must be sampled for every use, driving $\alpha \approx 1$ and collapsing effective data-access throughput, thereby pushing the system into a strongly entropy-limited regime.

Overall, the von Neumann architecture exposes a fundamental limitation: randomness is treated as an auxiliary resource rather than an integral component of data access. This structural mismatch between stochastic demand and entropy supply gives rise to the entropy bottleneck illustrated in Fig.~\ref{fig:bottleneck}, motivating alternative architectural paradigms.

\section*{\textcolor{sared}{\large Probabilistic Compute-in-Memory: Opportunities and Challenges}}

\noindent\textbf{From RNG bottleneck to probabilistic compute-in-memory}

\noindent The preceding section highlights a fundamental limitation of von Neumann architectures: as the probabilistic data ratio ($\alpha$) increases, system throughput becomes increasingly constrained by entropy generation, reducing effective data-access throughput under the unified model. This limitation arises from the physical separation between memory access and random number generation, which forces stochastic data to be generated, transported, and consumed through disjoint hardware pathways (Fig.~\ref{fig:bottleneck}(a)).

This observation motivates a complementary paradigm: \emph{probabilistic compute-in-memory} (p-CIM)~\cite{isscc_shot_noise}, illustrated in Fig.~\ref{fig:bottleneck}(b). The key idea is to embed entropy generation directly within the memory-access path, allowing stochastic sampling to occur in situ during data retrieval. By eliminating explicit random-number transport and aligning sampling with memory-array parallelism, pCIM increases effective data-access throughput by co-scaling entropy generation with memory bandwidth. This architectural shift is particularly beneficial in high-$\alpha$ regimes, where entropy supply dominates system performance.

\noindent\textbf{Entropy generation for probabilistic compute-in-memory}

\noindent From the unified data-access perspective, entropy sources should be evaluated not only by physical origin, but by their impact on effective data-access throughput, distribution programmability, and statistical fidelity, which together determine how efficiently stochastic sampling integrates with memory operations.

In conventional CMOS technologies, entropy generation relies on intrinsic noise and device variability. Dynamic sources exploit thermal noise~\cite{res_noise} and shot noise~\cite{isscc_shot_noise,JSSC_shot_noise}, captured through analog or time-domain sampling, while time-domain uncertainty in ring oscillators or delay lines~\cite{rint_osc} and metastability-based circuits~\cite{Metastability} provide alternative mechanisms. Static sources leverage device mismatch or leakage variation~\cite{pv1}. Although CMOS sources benefit from mature integration, their limited noise magnitude often requires amplification or post-processing, constraining entropy density and effective throughput in dense memory arrays.

Emerging devices provide stronger intrinsic stochasticity, including filament formation in resistive switching devices~\cite{Memristors_review}, phase transitions in phase-change memory~\cite{pcm}, probabilistic switching in spintronic devices~\cite{mram_rng}, ferroelectric polarization switching~\cite{fefet_cim}, and quantum tunneling in advanced CMOS~\cite{peiUncertaintyawareRoboticPerception2025a}. These mechanisms generate randomness directly from physical processes, enabling higher entropy density and scalability, and form the basis for embedding stochastic sampling within memory systems.

\noindent\textbf{Coupled parameter storage and sampling}

\noindent In \emph{coupled} probabilistic compute-in-memory architectures, parameter storage and entropy generation are integrated within the same physical device~\cite{Memristors_bayesian_review}. Stochastic behavior emerges directly during memory operations, enabling sampling without explicit random number generation. Because entropy generation is embedded in the data-access process, sampling throughput scales with memory-array parallelism, effectively increasing entropy throughput ($T_{\mathrm{ent}}$) and improving performance in high-$\alpha$ regimes.

This tight integration enables compact and energy-efficient implementations, particularly for workloads with frequent sampling. However, coupling storage and entropy generation introduces fundamental limitations. The statistical properties of generated samples are governed by device physics, which constrains distribution programmability and limits independent control of mean and variance. In many implementations, distribution parameters are entangled with device characteristics, restricting the achievable distribution space and often requiring hardware-aware training~\cite{Nature-communication}. In addition, write-based sampling mechanisms can introduce endurance concerns~\cite{NE-endurance}, while tightly coupled designs may struggle to support both deterministic and probabilistic modes within the same memory system.

\noindent\textbf{Decoupled parameter storage and sampling}

\noindent \emph{Decoupled} probabilistic compute-in-memory architectures separate deterministic parameter storage from entropy generation~\cite{STT-MRAM_cim}. A common formulation follows the reparameterization principle $x = \mu + \sigma \epsilon$, where $\mu$ and $\sigma$ are stored in memory and $\epsilon$ is generated by an entropy source. This separation enables independent control of distribution parameters, improving programmability and statistical fidelity.

From a system perspective, decoupled designs enhance distribution control but introduce additional overhead in entropy delivery. In \emph{near-memory} implementations, entropy is generated in peripheral circuits and must be transported or written back before computation, increasing data movement and reducing effective data-access throughput~\cite{STT-MRAM_cim}. In contrast, \emph{in-memory} entropy generation embeds stochastic behavior within the memory-access path~\cite{JSSC-Zephan}, aligning sampling throughput with array parallelism and improving efficiency. However, because both deterministic parameters and entropy-generation circuitry must coexist, decoupled designs generally incur higher hardware overhead than tightly coupled approaches.

\noindent\textbf{Architectural trade-offs under the unified model}

\noindent From the unified data-access perspective, probabilistic compute-in-memory architectures can be understood as modifying effective data-access throughput by co-designing entropy generation and memory access. Coupled designs maximize entropy throughput and are well-suited for entropy-dominated regimes (high $\alpha$), where sampling demand is high and throughput is critical. Decoupled designs provide improved distribution programmability and statistical fidelity, but at the cost of additional data movement and reduced effective throughput.

These trade-offs highlight a fundamental design space spanning efficiency, flexibility, and robustness, as illustrated in Figure.~\ref{fig:comparison}. No single architecture optimizes all three simultaneously, suggesting that future systems will require cross-layer co-design across devices, circuits, architectures, and algorithms. By integrating entropy generation into the memory-access pathway, probabilistic compute-in-memory offers a promising direction for overcoming the entropy bottleneck and enabling scalable probabilistic computation in trustworthy AI systems.

\begin{figurehere}
\vspace{3ex}
   \centering
    \includegraphics[width=0.95\linewidth]{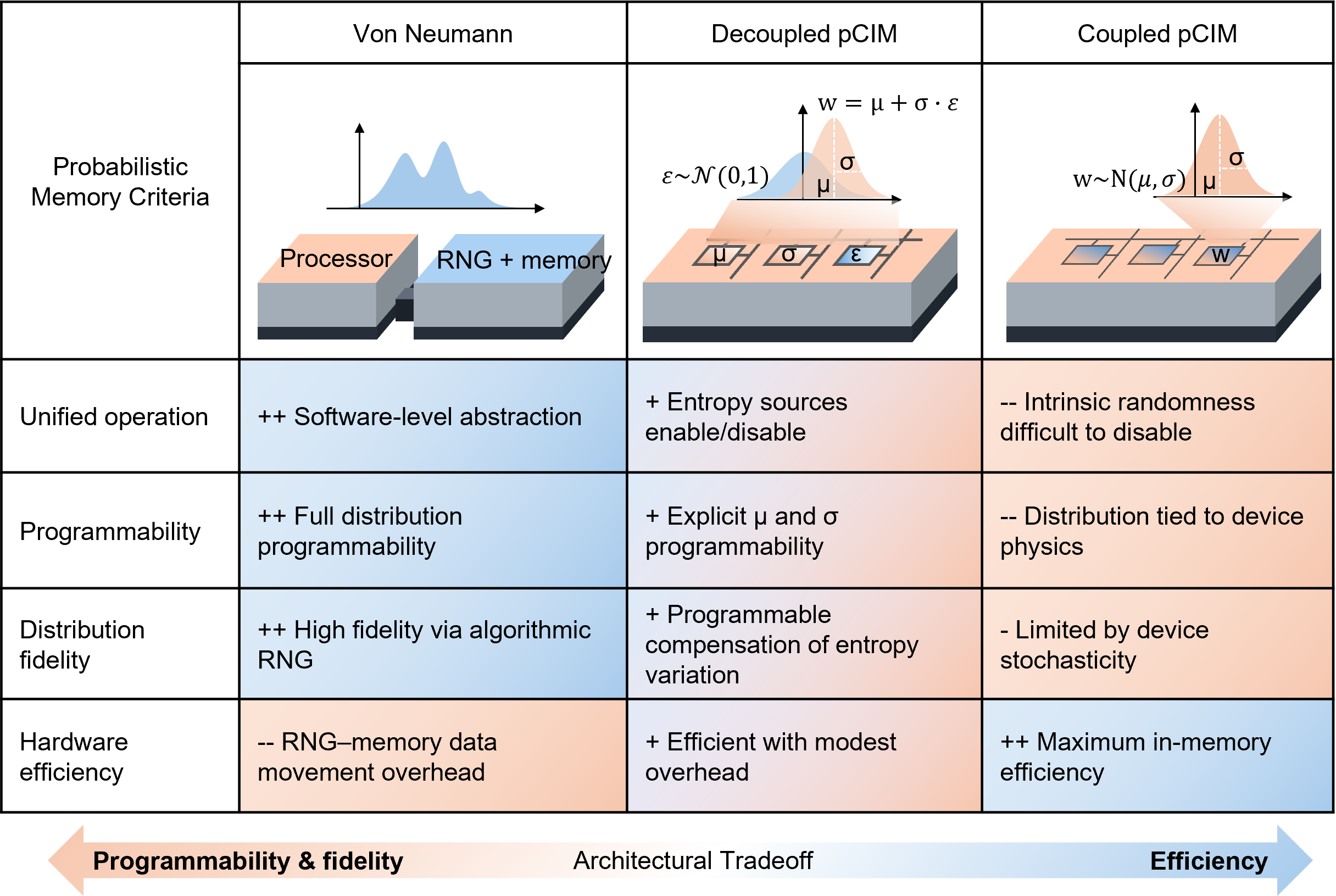}
    \captionsetup{parbox=none} % for caption to be split
    \caption{
\textbf{Architectural trade-offs in probabilistic compute-in-memory.}
\textit{Comparison of von Neumann, coupled pCIM, and decoupled pCIM across key criteria. Von Neumann architectures separate random number generation (RNG) from memory and compute, enabling high programmability and fidelity but incurring data-movement overhead. Coupled pCIM embeds entropy generation within memory, enabling in-situ sampling and high efficiency but limiting distribution control. Decoupled pCIM separates parameter storage and entropy generation (e.g., $w=\mu+\sigma\epsilon$), providing improved programmability and calibration with moderate efficiency. Together, these approaches define a continuum between programmability and hardware efficiency in probabilistic memory design.}
}
    \label{fig:comparison}
\end{figurehere}

\section*{\textcolor{sared}{\large Outlook}}

\noindent
Viewed through the unified data-access perspective, future memory architectures for trustworthy AI must balance efficiency, programmability, statistical fidelity, and system-level usability, with effective data-access throughput increasingly governed by entropy delivery as the probabilistic data ratio ($\alpha$) increases. Achieving scalable performance therefore requires co-optimizing entropy generation, distribution programmability, and statistical fidelity such that $T_{\mathrm{access}}$ scales with workload demand while managing device variability, minimizing overhead, and preserving seamless system integration.

\noindent\textbf{Technology scaling amplifies entropy—but not necessarily usable entropy.}
In deeply scaled CMOS, threshold voltage variation follows Pelgrom’s law~\cite{Pelgrom}, with $\sigma_{V_{th}} \propto 1/\sqrt{WL}$, while increased temperature amplifies $kT/C$ noise. Three-dimensional integration further introduces stochasticity from thermal gradients, interconnect coupling, and BEOL device process variations (Fig.~\ref{fig:outlook}(a)). 
In conventional systems, these effects are treated as non-idealities to be suppressed. In entropy-native architectures, they instead form distributed entropy reservoirs across the memory hierarchy. However, increased variability does not directly translate into higher effective entropy throughput. Spatial correlation, temporal drift, and aging can reduce usable entropy and degrade statistical fidelity, limiting improvements in $T_{\mathrm{access}}$ despite abundant physical randomness.

\noindent\textbf{From entropy generation to entropy shaping.}
Raw device noise rarely matches the distributions required by AI workloads. Circuit-level programmability—through bias modulation, transconductance control, analog accumulation, or embedded inverse-CDF approximations~\cite{vae_hdc_boyang}—enables mapping intrinsic variability to structured distributions such as Gaussian or mixture priors (Fig.~\ref{fig:outlook}(b)). Adjusting bias conditions controls variance through $g_m$ and capacitance scaling, while post-processing reshapes distribution tails and suppresses bias.
Entropy shaping defines a fundamental trade-off between programmability and throughput. Fine-grained distribution control typically introduces additional circuitry or processing stages, reducing effective data-access throughput. Efficient probabilistic memory systems must therefore balance distribution flexibility with entropy delivery, particularly in high-$\alpha$ regimes.

\noindent\textbf{Stochastic quality as a workload-dependent design dimension.}
Statistical requirements vary significantly across workloads, including distribution type, precision, correlation, and tail behavior. Bayesian neural networks emphasize accurate mean representation with relaxed variance precision~\cite{ICCAD}, generative models require long, decorrelated stochastic sequences~\cite{generative}, and Monte Carlo methods impose task-dependent constraints on tail accuracy and sampling efficiency~\cite{MC_meathod}.
These differences indicate that randomness is not a uniform resource. Statistical fidelity directly impacts how effectively entropy contributes to computation and thus influences effective throughput. Existing hardware metrics are often decoupled from algorithm-level performance, motivating workload-aware evaluation frameworks that connect device- and circuit-level stochastic properties to system-level outcomes (Fig.~\ref{fig:outlook}(c)).

\noindent\textbf{Probabilistic memory as a programmable system abstraction.}
Future systems will require architectural interfaces that treat entropy as a first-class computational resource (Fig.~\ref{fig:outlook}(d)), supported by instruction-set primitives for distribution-aware operations (e.g., \texttt{SAMPLE}, \texttt{READ\_DISTRIBUTION}, \texttt{SET\_VARIANCE}), compiler-level entropy scheduling, and probabilistic programming abstractions that enable direct control of stochastic behavior within memory. Such abstractions align software-level stochastic demand with hardware-level entropy delivery and enable cross-layer optimization, while complementary validation frameworks are needed to link device- and array-level entropy properties to system-level trust metrics, including calibration accuracy, robustness, and privacy guarantees. If deterministic memory defined the computing substrate of past decades, entropy-native memory architectures may define the foundation of trustworthy AI, requiring coordinated advances across devices, circuits, architectures, and software to transform variability into a scalable computational resource that directly contributes to effective data-access throughput.

\begin{figurehere}
   \centering
    \includegraphics[width=1\linewidth]{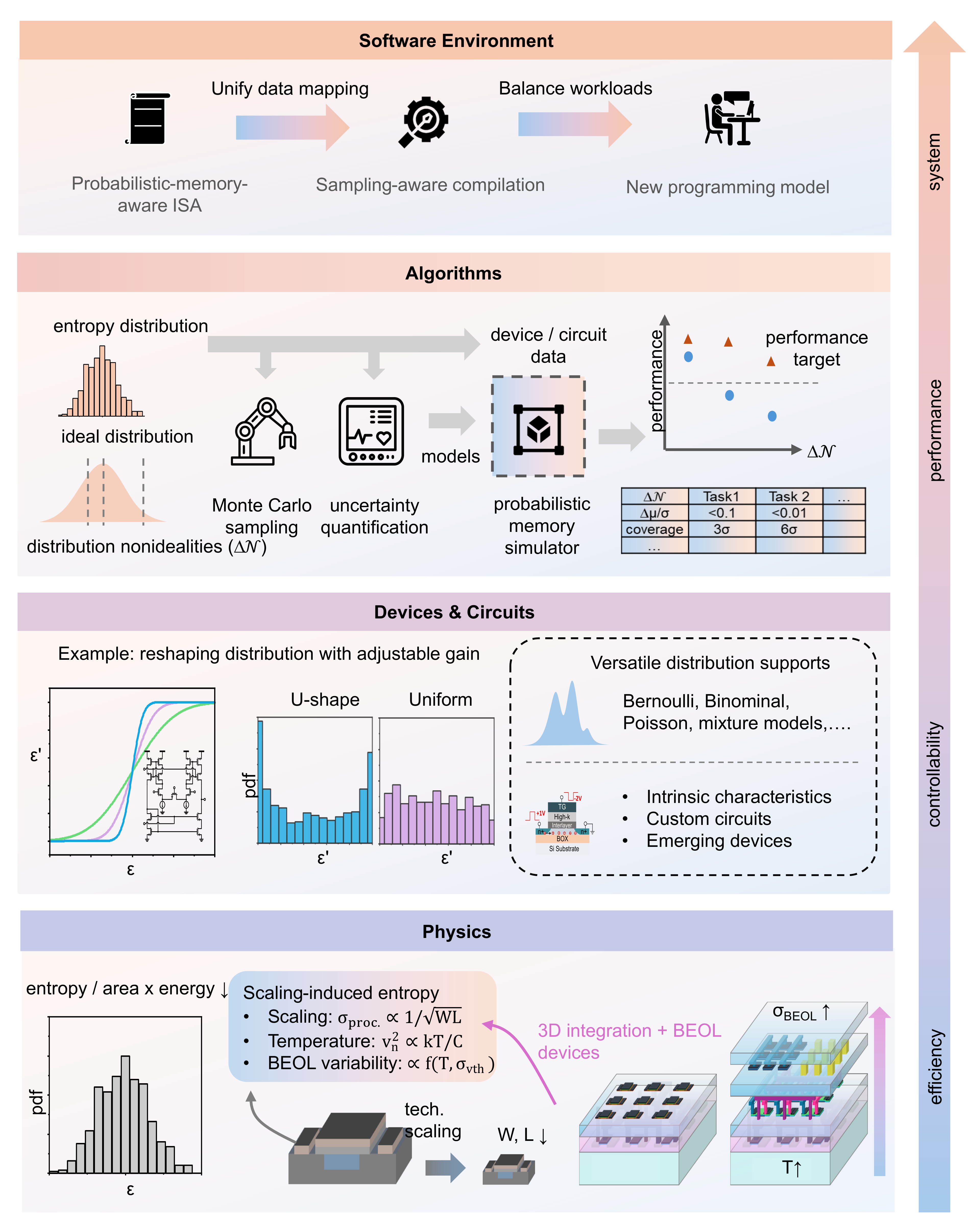}
    \captionsetup{parbox=none} % for caption to be split
\caption{
\textbf{Cross-layer design framework for probabilistic memory systems.}
\textit{\textbf{a,} At the physics level, intrinsic stochastic processes and scaling-induced variability determine the available entropy and its efficiency. 
\textbf{b,} At the device and circuit level, these entropy sources are harnessed and reshaped to realize different stochastic distributions through intrinsic device behavior, custom circuits, and emerging technologies. 
\textbf{c,} At the algorithm level, distribution non-idealities propagate to system performance through metrics such as uncertainty estimation and task accuracy, motivating the use of probabilistic memory simulators\cite{pm_dac} for hardware–algorithm co-design. 
\textbf{d,} At the software level, new abstractions—including probabilistic instructions, sampling-aware compilation, and workload balancing—enable integration of stochastic operations into programming models. Together, these layers highlight the need for coordinated optimization of entropy generation, distribution programmability, and system efficiency in probabilistic memory architectures.}
}
    \label{fig:outlook}
\end{figurehere}

\section*{\textcolor{sared}{\large Conclusion}} 

\noindent In summary, this Perspective establishes a unified data-access framework that treats deterministic and probabilistic computation within a common abstraction, revealing entropy delivery as a fundamental system-level constraint alongside memory bandwidth. By introducing the probabilistic data ratio and a roofline-like model, we show how modern AI workloads increasingly transition from data-bound to entropy-bound regimes, motivating memory architectures that integrate stochastic sampling as a first-class primitive. Addressing this shift requires co-design across devices, circuits, architectures, and software to ensure efficient, programmable, and statistically robust entropy generation. Such entropy-native memory systems provide a pathway toward scalable, trustworthy AI, where variability is no longer a limitation but a resource that directly contributes to computation.

\section*{\large Data availability}
This perspective paper does not contain any new experimental data. All data discussed are available from the cited literature.

\bibliography{ref}

@inproceedings{rezende2014stochastic,
  title={Stochastic backpropagation and approximate inference in deep generative models},
  author={Rezende, Danilo Jimenez and Mohamed, Shakir and Wierstra, Daan},
  booktitle={International conference on machine learning},
  pages={1278--1286},
  year={2014}
}

@inproceedings{bernstein2008chacha,
  title={ChaCha, a variant of Salsa20},
  author={Bernstein, Daniel J and others},
  booktitle={Workshop record of SASC},
  volume={8},
  number={1},
  pages={3--5},
  year={2008},
  organization={Lausanne, Switzerland}
}

@article{vae,
  title={Auto-encoding variational bayes},
  author={Kingma, Diederik P and Welling, Max},
  journal={arXiv preprint arXiv:1312.6114},
  year={2013}
}

@Article{trng_throughput,
author = {Cao, Yuan and Liu, Wanyi and Qin, Lan and Liu, Bingqiang and Chen, Shuai and Ye, Jing and Xia, Xianzhao and Wang, Chao},
title = {Entropy Sources Based on Silicon Chips: True Random Number Generator and Physical Unclonable Function},
journal = {Entropy},
volume = {24},
number = {11},
article-number = {1566},
year = {2022},
PubMedID = {36359655},
ISSN = {1099-4300},
DOI = {10.3390/e24111566}
}

@article{silver_mastering_2017,
	title = {Mastering the game of {Go} without human knowledge},
	volume = {550},
	issn = {1476-4687},
	doi = {10.1038/nature24270},
	number = {7676},
	journal = {Nature},
	author = {Silver, David and Schrittwieser, Julian and Simonyan, Karen and Antonoglou, Ioannis and Huang, Aja and Guez, Arthur and Hubert, Thomas and Baker, Lucas and Lai, Matthew and Bolton, Adrian and Chen, Yutian and Lillicrap, Timothy and Hui, Fan and Sifre, Laurent and van den Driessche, George and Graepel, Thore and Hassabis, Demis},
	month = oct,
	year = {2017},
	pages = {354--359},
}

@inproceedings{jun2014memory,
  title={Memory (and time) efficient sequential Monte Carlo},
  author={Jun, Seong-Hwan and Bouchard-C{\^o}t{\'e}, Alexandre},
  booktitle={International Conference on Machine Learning},
  pages={514--522},
  year={2014},
  organization={PMLR}
}

@inproceedings{vae_hdc_boyang,
author = {Cheng, Boyang and Liu, Jianbo and Davis, Steven and Enciso, Zephan M. and Zhang, Yiyang and Cao, Ningyuan},
title = {VAE-HDC: Efficient and Secure Hyper-dimensional Encoder Leveraging Variation Analog Entropy},
year = {2024},
isbn = {9798400706011},
publisher = {Association for Computing Machinery},
address = {New York, NY, USA},
doi = {10.1145/3649329.3658486},
booktitle = {Proceedings of the 61st ACM/IEEE Design Automation Conference},
articleno = {307},
numpages = {6},
location = {San Francisco, CA, USA},
series = {DAC '24}
}

@Article{particle_filtering,
author={Abd El-Halym, Howida A.
and Mahmoud, Imbaby Ismail
and Habib, S. E. D.},
title={Proposed hardware architectures of particle filter for object tracking},
journal={EURASIP Journal on Advances in Signal Processing},
year={2012},
month={Jan},
day={23},
volume={2012},
number={1},
pages={17},
issn={1687-6180},
doi={10.1186/1687-6180-2012-17},
}

@article{mc_tree_search,
author = {Fan, Shaoze and Zhang, Shun and Liu, Jianbo and Cao, Ningyuan and Guo, Xiaoxiao and Li, Jing and Zhang, Xin},
title = {Power Converter Circuit Design Automation Using Parallel Monte Carlo Tree Search},
year = {2022},
issue_date = {March 2023},
publisher = {Association for Computing Machinery},
address = {New York, NY, USA},
volume = {28},
number = {2},
issn = {1084-4309},
doi = {10.1145/3549538},
journal = {ACM Trans. Des. Autom. Electron. Syst.},
month = dec,
articleno = {17},
numpages = {33}
}

@article{cim_nature_sun_2023,
	title = {A full spectrum of computing-in-memory technologies},
	volume = {6},
	issn = {2520-1131},
	doi = {10.1038/s41928-023-01053-4},
	number = {11},
	journal = {Nature Electronics},
	author = {Sun, Zhong and Kvatinsky, Shahar and Si, Xin and Mehonic, Adnan and Cai, Yimao and Huang, Ru},
	month = nov,
	year = {2023},
	pages = {823--835},
}

@article{memory_wall_wulf,
author = {Wulf, Wm. A. and McKee, Sally A.},
title = {Hitting the memory wall: implications of the obvious},
year = {1995},
issue_date = {March 1995},
publisher = {Association for Computing Machinery},
address = {New York, NY, USA},
volume = {23},
number = {1},
issn = {0163-5964},
doi = {10.1145/216585.216588},
journal = {SIGARCH Comput. Archit. News},
month = mar,
pages = {20–24},
numpages = {5}
}

@article{google_tpu,
author = {Jouppi, Norman P. and Young, Cliff and Patil, Nishant and Patterson, David and Agrawal, Gaurav and Bajwa, Raminder and Bates, Sarah and Bhatia, Suresh and Boden, Nan and Borchers, Al and Boyle, Rick and Cantin, Pierre-luc and Chao, Clifford and Clark, Chris and Coriell, Jeremy and Daley, Mike and Dau, Matt and Dean, Jeffrey and Gelb, Ben and Ghaemmaghami, Tara Vazir and Gottipati, Rajendra and Gulland, William and Hagmann, Robert and Ho, C. Richard and Hogberg, Doug and Hu, John and Hundt, Robert and Hurt, Dan and Ibarz, Julian and Jaffey, Aaron and Jaworski, Alek and Kaplan, Alexander and Khaitan, Harshit and Killebrew, Daniel and Koch, Andy and Kumar, Naveen and Lacy, Steve and Laudon, James and Law, James and Le, Diemthu and Leary, Chris and Liu, Zhuyuan and Lucke, Kyle and Lundin, Alan and MacKean, Gordon and Maggiore, Adriana and Mahony, Maire and Miller, Kieran and Nagarajan, Rahul and Narayanaswami, Ravi and Ni, Ray and Nix, Kathy and Norrie, Thomas and Omernick, Mark and Penukonda, Narayana and Phelps, Andy and Ross, Jonathan and Ross, Matt and Salek, Amir and Samadiani, Emad and Severn, Chris and Sizikov, Gregory and Snelham, Matthew and Souter, Jed and Steinberg, Dan and Swing, Andy and Tan, Mercedes and Thorson, Gregory and Tian, Bo and Toma, Horia and Tuttle, Erick and Vasudevan, Vijay and Walter, Richard and Wang, Walter and Wilcox, Eric and Yoon, Doe Hyun},
title = {In-Datacenter Performance Analysis of a Tensor Processing Unit},
year = {2017},
issue_date = {May 2017},
publisher = {Association for Computing Machinery},
address = {New York, NY, USA},
volume = {45},
number = {2},
issn = {0163-5964},
doi = {10.1145/3140659.3080246},
journal = {SIGARCH Comput. Archit. News},
month = jun,
pages = {1–12},
numpages = {12}
}

@article{wang_safe_2024,
	title = {Safe, secure and trustworthy compute-in-memory accelerators},
    author = {Wang, Ziyu and Wu, Yuting and Park, Yongmo and Lu, Wei D.},
    journal = {Nature Electronics},
	volume = {7},
    number = {12},
    pages = {1086--1097},
    year = {2024},
	issn = {2520-1131},
	doi = {10.1038/s41928-024-01312-y},
	journaltitle = {Nature Electronics},
	shortjournal = {Nature Electronics},
	date = {2024-12-01},
}

@article{ghahramani_probabilistic_2015,
	title = {Probabilistic machine learning and artificial intelligence},
    author = {Ghahramani, Zoubin},
    journal = {Nature},
	volume = {521},
    number = {7553},
	pages = {452--459},
	  year = {2015},
}

@article{automated_driving_nn,
author = {Nathan A. Spielberg  and Matthew Brown  and Nitin R. Kapania  and John C. Kegelman  and J. Christian Gerdes },
title = {Neural network vehicle models for high-performance automated driving},
journal = {Science Robotics},
volume = {4},
number = {28},
pages = {eaaw1975},
year = {2019},
doi = {10.1126/scirobotics.aaw1975}}

@article{science_robotics_uhang,
author = {Yuhang Hu  and Jiong Lin  and Judah Allen Goldfeder  and Philippe M. Wyder  and Yifeng Cao  and Steven Tian  and Yunzhe Wang  and Jingran Wang  and Mengmeng Wang  and Jie Zeng  and Cameron Mehlman  and Yingke Wang  and Delin Zeng  and Boyuan Chen  and Hod Lipson },
title = {Learning realistic lip motions for humanoid face robots},
journal = {Science Robotics},
volume = {11},
number = {110},
pages = {eadx3017},
year = {2026},
doi = {10.1126/scirobotics.adx3017}}

@inproceedings{ribeiro_lime_2016,
author = {Ribeiro, Marco Tulio and Singh, Sameer and Guestrin, Carlos},
title = {Why Should I Trust You?: Explaining the Predictions of Any Classifier},
year = {2016},
isbn = {9781450342322},
publisher = {Association for Computing Machinery},
address = {New York, NY, USA},
doi = {10.1145/2939672.2939778},
booktitle = {Proceedings of the 22nd ACM SIGKDD International Conference on Knowledge Discovery and Data Mining},
pages = {1135–1144},
numpages = {10},
location = {San Francisco, California, USA},
series = {KDD '16}
}

@INPROCEEDINGS{ICCAD,
  author={Pei, Likai and Qin, Yifan and Enciso, Zephan M. and Cheng, Boyang and Liu, Jianbo and Davis, Steven and Jia, Zhenge and Niemier, Michael and Shi, Yiyu and Hu, X. Sharon and Cao, Ningyuan },
  booktitle={2024 IEEE/ACM International Conference On Computer Aided Design (ICCAD)}, 
  title={Towards Uncertainty-Quantifiable Biomedical Intelligence: Mixed-signal Compute-in-Entropy for Bayesian Neural Networks}, 
  year={2024},
  volume={},
  number={},
  pages={},
  doi={}}

@article{Nature-communication,
  title={Bringing uncertainty quantification to the extreme-edge with memristor-based Bayesian neural networks},
  author={Bonnet, Djohan and Hirtzlin, Tifenn and Majumdar, Atreya and Dalgaty, Thomas and Esmanhotto, Eduardo and Meli, Valentina and Castellani, Niccolo and Martin, Simon and Nodin, Jean-Fran{\c{c}}ois and Bourgeois, Guillaume and others},
  journal={Nature Communications},
  volume={14},
  number={1},
  pages={7530},
  year={2023},
  publisher={Nature Publishing Group UK London}
}

@article{metwallyPredictionMetabolicSubphenotypes2025,
  title = {Prediction of Metabolic Subphenotypes of Type 2 Diabetes via Continuous Glucose Monitoring and Machine Learning},
  author = {Metwally, Ahmed A. and Perelman, Dalia and Park, Heyjun and Wu, Yue and Jha, Alokkumar and Sharp, Seth and Celli, Alessandra and Ayhan, Ekrem and Abbasi, Fahim and Gloyn, Anna L. and McLaughlin, Tracey and Snyder, Michael P.},
  year = 2025,
  month = aug,
  journal = {Nature Biomedical Engineering},
  volume = {9},
  number = {8},
  pages = {1222--1239},
  publisher = {Nature Publishing Group},
  issn = {2157-846X},
  doi = {10.1038/s41551-024-01311-6},
  urldate = {2025-11-14},
  copyright = {2024 The Author(s)},
  langid = {english}
}

@article{nutiExplainableBayesianDecision2021,
  title = {An {{Explainable Bayesian Decision Tree Algorithm}}},
  author = {Nuti, Giuseppe and Jim{\'e}nez Rugama, Llu{\'i}s Antoni and Cross, Andreea-Ingrid},
  year = 2021,
  month = mar,
  journal = {Frontiers in Applied Mathematics and Statistics},
  volume = {7},
  publisher = {Frontiers},
  issn = {2297-4687},
  doi = {10.3389/fams.2021.598833},
  urldate = {2025-11-16},
  langid = {english}
}

@article{davisInSituPrivacyMixedSignal2024,
  title = {In-{{Situ Privacy}} via {{Mixed-Signal Perturbation}} and {{Hardware-Secure Data Reversibility}}},
  author = {Davis, Steven and Liu, Jianbo and Cheng, Boyang and Chang, Muya and Cao, Ningyuan},
  year = 2024,
  month = jun,
  journal = {IEEE Transactions on Circuits and Systems I: Regular Papers},
  volume = {71},
  number = {6},
  pages = {2538--2549},
  issn = {1558-0806},
  doi = {10.1109/TCSI.2024.3383337},
  urldate = {2025-11-16}
}

@inproceedings{peiUncertaintyawareRoboticPerception2025a,
  title = {Towards {{Uncertainty-aware Robotic Perception}} via {{Mixed-signal BNN Engine Leveraging Probabilistic Quantum Tunneling}}},
  booktitle = {2025 62nd {{ACM}}/{{IEEE Design Automation Conference}} ({{DAC}})},
  author = {Pei, Likai and Zhou, Yu and Wang, Xingtian and Zhao, Xueji and Huang, Wanxin and Cheng, Boyang and Mulaosmanovic, Halid and Duenkel, Stefan and Kleimaier, Dominik and Beyer, Sven and Ni, Kai and Hou, Mengxue and Niemier, Michael and Cao, Ningyuan},
  year = 2025,
  month = jun,
  pages = {1--7},
  doi = {10.1109/DAC63849.2025.11132881},
  urldate = {2025-11-16}
}

@ARTICLE{res_noise,
  author={Petrie, C.S. and Connelly, J.A.},
  journal={IEEE Transactions on Circuits and Systems I: Fundamental Theory and Applications}, 
  title={A noise-based IC random number generator for applications in cryptography}, 
  year={2000},
  volume={47},
  number={5},
  pages={615-621},
  doi={10.1109/81.847868}}

@INPROCEEDINGS{isscc_shot_noise,
  author={Liu, Jianbo and Enciso, Zephan and Cheng, Boyang and Pei, Likai and Davis, Steven and Qin, Yifan and Jia, Zhenge and Hu, Xiaobo Sharon and Shi, Yiyu and Cao, Ningyuan},
  booktitle={2025 IEEE International Solid-State Circuits Conference (ISSCC)}, 
  title={A 65nm Uncertainty-Quantifiable Ventricular Arrhythmia Detection Engine with $\mathbf{1.75}\boldsymbol{\mu}\mathbf{J}$ Per Inference}, 
  year={2025},
  volume={68},
  number={},
  pages={1-3},
  doi={10.1109/ISSCC49661.2025.10904610}}

@article{JSSC_shot_noise,
  title={In-memory unified TRNG and multi-bit PUF for ubiquitous hardware security},
  author={Taneja, Sachin and Rajanna, Viveka Konandur and Alioto, Massimo},
  journal={IEEE Journal of Solid-State Circuits},
  volume={57},
  number={1},
  pages={153--166},
  year={2021},
  publisher={IEEE}
}

@article{rint_osc,
  title={A provably secure true random number generator with built-in tolerance to active attacks},
  author={Sunar, Berk and Martin, William J and Stinson, Douglas R},
  journal={IEEE Transactions on computers},
  volume={56},
  number={1},
  pages={109--119},
  year={2007},
  publisher={IEEE}
}

@inproceedings{Metastability,
  title={Design and implementation of a true random number generator based on digital circuit artifacts},
  author={Epstein, Michael and Hars, Laszlo and Krasinski, Raymond and Rosner, Martin and Zheng, Hao},
  booktitle={International Workshop on Cryptographic Hardware and Embedded Systems},
  pages={152--165},
  year={2003},
  organization={Springer}
}

@inproceedings{pv1,
  title={8.3 A 553F 2 2-transistor amplifier-based Physically Unclonable Function (PUF) with 1.67\% native instability},
  author={Yang, Kaiyuan and Dong, Qing and Blaauw, David and Sylvester, Dennis},
  booktitle={2017 IEEE International Solid-State Circuits Conference (ISSCC)},
  pages={146--147},
  year={2017},
  organization={IEEE}
}

@INPROCEEDINGS{mram_rng,
  author={Yang, Kaiyuan and Dong, Qing and Wang, Zhehong and Shih, Yi-Chun and Chih, Yu-Der and Chang, Jonathan and Blaauw, David and Svlvester, Dennis},
  booktitle={2018 IEEE Symposium on VLSI Circuits}, 
  title={A 28NM Integrated True Random Number Generator Harvesting Entropy from MRAM}, 
  year={2018},
  volume={},
  number={},
  pages={171-172},
  doi={10.1109/VLSIC.2018.8502431}}

@article{Memristors_bayesian_review,
  title={Bayesian electronics for trustworthy artificial intelligence},
  author={Querlioz, Damien and Vianello, Elisa},
  journal={Nature Reviews Electrical Engineering},
  pages={1--10},
  year={2025},
  publisher={Nature Publishing Group UK London}
}

@article{Memristors_review,
  title={Memristors for Bayesian in-memory computing},
  author={Dalgaty, Thomas and Vianello, Elisa and Querlioz, Damien},
  journal={Nature Materials},
  pages={1--4},
  year={2025},
  publisher={Nature Publishing Group UK London}
}

@article{pcm,
  title={Transforming memristor noises into computational innovations},
  author={Ding, Chenchen and Ren, Yuan and Liu, Zhengwu and Wong, Ngai},
  journal={Communications Materials},
  volume={6},
  number={1},
  pages={149},
  year={2025},
  publisher={Nature Publishing Group UK London}
}

@article{fefet_cim,
  title={Ferroelectric NAND for efficient hardware bayesian neural networks},
  author={Song, Minsuk and Koo, Ryun-Han and Kim, Jangsaeng and Han, Chang-Hyeon and Yim, Jiyong and Ko, Jonghyun and Yoo, Sijung and Choe, Duk-hyun and Kim, Sangwook and Shin, Wonjun and others},
  journal={Nature Communications},
  volume={16},
  number={1},
  pages={6879},
  year={2025},
  publisher={Nature Publishing Group UK London}
}

@book{knuth1997art,
  author    = {Donald E. Knuth},
  title     = {The Art of Computer Programming, Volume 2: Seminumerical Algorithms},
  publisher = {Addison-Wesley},
  year      = {1997},
  edition   = {3rd}
}

@article{matsumotoMersenneTwister623dimensionally1998,
author = {Matsumoto, Makoto and Nishimura, Takuji},
title = {Mersenne twister: a 623-dimensionally equidistributed uniform pseudo-random number generator},
year = {1998},
issue_date = {Jan. 1998},
publisher = {Association for Computing Machinery},
address = {New York, NY, USA},
volume = {8},
number = {1},
issn = {1049-3301},
doi = {10.1145/272991.272995},
journal = {ACM Trans. Model. Comput. Simul.},
month = jan,
pages = {3–30},
numpages = {28}
}

@article{malikGaussianRandomNumber2016,
  title = {Gaussian {{Random Number Generation}}: {{A Survey}} on {{Hardware Architectures}}},
  shorttitle = {Gaussian {{Random Number Generation}}},
  author = {Malik, Jamshaid Sarwar and Hemani, Ahmed},
  year = 2016,
  month = nov,
  journal = {ACM Comput. Surv.},
  volume = {49},
  number = {3},
  pages = {53:1--53:37},
  issn = {0360-0300},
  doi = {10.1145/2980052},
  urldate = {2026-01-03},
}

@article{thomasGaussianRandomNumber2007a,
  title = {Gaussian Random Number Generators},
  author = {Thomas, David B. and Luk, Wayne and Leong, Philip H.W. and Villasenor, John D.},
  year = 2007,
  month = nov,
  journal = {ACM Comput. Surv.},
  volume = {39},
  number = {4},
  pages = {11--es},
  issn = {0360-0300},
  doi = {10.1145/1287620.1287622},
  urldate = {2026-02-06},
}

@inproceedings{khwa14216nm216kb2025,
  title = {14.2 {{A}} 16nm 216kb, 188.{{4TOPS}}/{{W}} and 133.{{5TFLOPS}}/{{W Microscaling Multi-Mode Gain-Cell CIM Macro Edge-AI Devices}}},
  booktitle = {2025 {{IEEE International Solid-State Circuits Conference}} ({{ISSCC}})},
  author = {Khwa, Win-San and Wu, Ping-Chun and Su, Jian-Wei and Cheng, Chiao-Yen and Hsu, Jun-Ming and Chen, Yu-Chen and Hsieh, Le-Jung and Bai, Jyun-Cheng and Kao, Yu-Sheng and Lou, Tsung-Han and Lele, Ashwin Sanjay and Wu, Jui-Jen and Tien, Jen-Chun and Lo, Chung-Chuan and Liu, Ren-Shuo and Hsieh, Chih-Cheng and Tang, Kea-Tiong and Chang, Meng-Fan},
  year = 2025,
  month = feb,
  volume = {68},
  pages = {1--3},
  issn = {2376-8606},
  doi = {10.1109/ISSCC49661.2025.10904606},
  urldate = {2026-02-06},
}

@article{writing_reading_application,
  title={Deep Bayesian active learning using in-memory computing hardware},
  author={Lin, Yudeng and Gao, Bin and Tang, Jianshi and Zhang, Qingtian and Qian, He and Wu, Huaqiang},
  journal={Nature Computational Science},
  volume={5},
  number={1},
  pages={27--36},
  year={2025},
  publisher={Nature Publishing Group US New York}
}

@INPROCEEDINGS{STT-MRAM_cim,
  title = {14.1 {{A}} 22nm 104.{{5TOPS}}/{{W}} \textmu -{{NMC-$\Delta$-IMC Heterogeneous STT-MRAM CIM Macro}} for {{Noise-Tolerant Bayesian Neural Networks}}},
  booktitle = {2025 {{IEEE International Solid-State Circuits Conference}} ({{ISSCC}})},
  author = {You, De-Qi and Khwa, Win-San and Zhang, Bo and Chen, Fang-Yi and Lee, Andrew and Hung, Yu-Cheng and Li, Yi-Ming and Wang, Yu-Hui and Lo, Chung-Chuan and Liu, Ren-Shuo and Tang, Kea-Tiong and Hsieh, Chih-Cheng and Chih, Yu-Der and Chang, Tsung-Yung Jonathan and Chang, Meng-Fan},
  year = 2025,
  month = feb,
  volume = {68},
  pages = {1--3},
  issn = {2376-8606},
  doi = {10.1109/ISSCC49661.2025.10904540},
  urldate = {2026-02-09},
}

@article{distribution_conversion,
  title={Probabilistic neural computing with stochastic devices},
  author={Misra, Shashank and Bland, Leslie C and Cardwell, Suma G and Incorvia, Jean Anne C and James, Conrad D and Kent, Andrew D and Schuman, Catherine D and Smith, J Darby and Aimone, James B},
  journal={Advanced Materials},
  volume={35},
  number={37},
  pages={2204569},
  year={2023},
  publisher={Wiley Online Library}
}

@misc{cao_survey_2023,
    title = {A survey on generative diffusion model},
    author = {Cao, Hanqun and Tan, Cheng and Gao, Zhangyang and Xu, Yilun and Chen, Guangyong and Heng, Pheng-Ann and Li, Stan Z.},
    doi = {10.48550/arXiv.2209.02646},
    urldate = {2025-07-20},
    publisher = {arXiv},
    month = dec,
    year = {2023},
    note = {arXiv:2209.02646 [cs]},
}

@article{NE-endurance,
  title={In situ learning using intrinsic memristor variability via Markov chain Monte Carlo sampling},
  author={Dalgaty, Thomas and Castellani, Niccolo and Turck, Cl{\'e}ment and Harabi, Kamel-Eddine and Querlioz, Damien and Vianello, Elisa},
  journal={Nature Electronics},
  volume={4},
  number={2},
  pages={151--161},
  year={2021},
  publisher={Nature Publishing Group UK London}
}

@ARTICLE{JSSC-Zephan,
  author={Enciso, Zephan M. and Liu, Jianbo and Cheng, Boyang and Pei, Likai and Davis, Steven and Qin, Yifan and Jia, Zhenge and Hu, Xiaobo Sharon and Shi, Yiyu and Niemier, Michael and Cao, Ningyuan},
  journal={IEEE Journal of Solid-State Circuits}, 
  title={A 350-pW Implantable Ventricular Arrhythmia Detection Engine With Bayesian Uncertainty Quantification in 65-nm CMOS}, 
  year={2026},
  volume={},
  number={},
  pages={1-11},
  doi={10.1109/JSSC.2026.3669040}}

@ARTICLE{Pelgrom,
  author={Pelgrom, M.J.M. and Duinmaijer, A.C.J. and Welbers, A.P.G.},
  journal={IEEE Journal of Solid-State Circuits}, 
  title={Matching properties of MOS transistors}, 
  year={1989},
  volume={24},
  number={5},
  pages={1433-1439},
  doi={10.1109/JSSC.1989.572629}}

@inproceedings{generative,
author = {Karras, Tero and Aittala, Miika and Laine, Samuli and Aila, Timo},
title = {Elucidating the design space of diffusion-based generative models},
year = {2022},
isbn = {9781713871088},
publisher = {Curran Associates Inc.},
address = {Red Hook, NY, USA},
booktitle = {Proceedings of the 36th International Conference on Neural Information Processing Systems},
articleno = {1926},
numpages = {13},
location = {New Orleans, LA, USA},
series = {NIPS '22}
}

@article{MC_meathod,
author = {Park, S. K. and Miller, K. W.},
title = {Random number generators: good ones are hard to find},
year = {1988},
issue_date = {Oct. 1988},
publisher = {Association for Computing Machinery},
address = {New York, NY, USA},
volume = {31},
number = {10},
issn = {0001-0782},
doi = {10.1145/63039.63042},
journal = {Commun. ACM},
month = oct,
pages = {1192–1201},
numpages = {10}
}

@INPROCEEDINGS{pm_dac,
  author={Pei, Likai and Zheng, Jiahao and Zhao, Xueji and Ye, Emilie and Liu, Jianbo and Tao, Hanqing and Lee, Ming-Yen and Qin, Ruiyang and Shi, Yiyu and Yu, Shimeng and X.,Sharon, Hu  and Cao, Ningyuan},
  booktitle={2026 63rd ACM/IEEE Design Automation Conference (DAC)}, 
  title={Probabilistic Memory Design for Efficient Trustworthy Edge Intelligence}, 
  year={2026},
  volume={},
  number={},
  pages={1-7},
  doi={}}

@inproceedings{luoBenchmarking2024,
  title = {Benchmarking and {{Dissecting}} the {{Nvidia Hopper GPU Architecture}}},
  booktitle = {2024 {{IEEE International Parallel}} and {{Distributed Processing Symposium}} ({{IPDPS}})},
  author = {Luo, Weile and Fan, Ruibo and Li, Zeyu and Du, Dayou and Wang, Qiang and Chu, Xiaowen},
  year = 2024,
  month = may,
  pages = {656--667},
  issn = {1530-2075},
  doi = {10.1109/IPDPS57955.2024.00064},
  urldate = {2026-03-21}
}
\bibliographystyle{naturemag}
\section*{\large Acknowledgments}

This work was primarily supported by NSF 2426639, 2346953, 2347024, and 2404874.

\section*{\large Author contributions}

X.Z. and L.P. led the manuscript preparation.
J.L. provides reviews on GRNG, SRAM and CIM throughput efficiency.
N.C. and K.N. led the project.
All authors contributed to the discussions.
X.Z. and L.P. contributed equally to this work.

\section*{\large Competing interests}
The authors declare that they have no competing interests.

\end{document}